\pdfoutput=1

\documentclass[11pt]{article}

\usepackage{ACL2023}

\usepackage{times}
\usepackage{latexsym}
\usepackage{booktabs}
\usepackage{multirow}
\usepackage{enumerate}
\usepackage{xcolor}
\usepackage{graphicx}
\usepackage{amsmath}
\usepackage{listings}
\usepackage{xcolor}
\usepackage{float}
\usepackage{xspace}

\usepackage[T1]{fontenc}

\usepackage[utf8]{inputenc}

\usepackage{microtype}

\usepackage{inconsolata}

\title{JIANG: Chinese Open Foundation Language Model}

\author{Qinhua Duan$^1$, Wenchao Gu, Yujia Chen, Wenxin Mao, Zewen Tian, Hui Cao \\ 
    $^1$ KDF, China.\\	 \\
}

\begin{document}
\maketitle
\begin{abstract}

With the advancements in large language model technology, it has showcased capabilities that come close to those of human beings across various tasks. This achievement has garnered significant interest from companies and scientific research institutions, leading to substantial investments in the research and development of these models. While numerous large models have emerged during this period, the majority of them have been trained primarily on English data. Although they exhibit decent performance in other languages, such as Chinese, their potential remains limited due to factors like vocabulary design and training corpus. Consequently, their ability to fully express their capabilities in Chinese falls short. To address this issue, we introduce the model named JIANG (Chinese pinyin of ginger) specifically designed for the Chinese language. We have gathered a substantial amount of Chinese corpus to train the model and have also optimized its structure. The extensive experimental results demonstrate the excellent performance of our model.
\end{abstract}

\section{Introduction}
Large-scale language models (LLMs) have made significant progress in natural language processing and artificial intelligence, resulting in remarkable advancements in these fields. These models are trained with vast amounts of textual data and exhibit exceptional performance in various natural language processing tasks, such as language translation, text summarization, and question answering. One notable example is the ChatGPT model, which has demonstrated its potential in a wide range of domains, including education, healthcare, reasoning, text generation, human-computer interaction, and scientific research. However, applying these models to the natural language processing tasks in Chinese presents challenges due to their predominantly English-oriented training corpora.

The existing large-scale models have shown inferior performance in the Chinese domain, emphasizing the urgent necessity for superior models specifically designed for Chinese language to bridge this gap. Concurrently, there is a growing demand for research on Chinese pre-training base models in today's global landscape. Chinese, being one of the most widely spoken languages globally, possesses a vast user base and a rich cultural heritage. However, current large-scale models predominantly focus on the English language, paying relatively less attention to Chinese. To fulfill the increasing need for Chinese language processing, it is crucial to conduct further research on Chinese pre-training base models in order to enhance the capabilities and expressive power of Chinese models. This effort will contribute to advancing Chinese language processing technology and enhancing language services and experiences for Chinese users.

This paper presents JIANG, an open-source large-scale Chinese model designed to address the specific needs of the Chinese language. Unlike current mainstream models in the industry, like LLaMA, Pythia (formerly known as GPT-NeoX), Cerebras, and StableLM, which are predominantly trained on English, JIANG focuses on leveraging Chinese language capabilities. However, these existing models, despite possessing some Chinese language skills, have limited performance in the Chinese domain due to factors such as vocabulary design and training corpora. As a result, they are unable to fully unlock the potential of Chinese models.

It is important to note that the development plan for JIANG was established prior to the launch of ChatGLM and MOSS. The training of JIANG begins in mid-April, between the releases dates of ChatGLM and MOSS. However, as of now, no technical reports or papers have been published about ChatGLM and MOSS.

Our objective is to actively contribute to the research and development of Chinese large-scale models through the open-source release of JIANG. We firmly believe that by fostering sharing and collaboration, we can attract talented individuals to drive advancements in the field. These dedicated researchers will bring fresh perspectives, innovative techniques, and advanced methodologies, infusing Chinese large-scale models with vitality. We aim to deliver precise and intelligent language processing services to Chinese users. Let us elevate the research and application of Chinese large-scale models to unprecedented levels.

\section{Approach}

We follow the previous work~\cite{abs-2302-13971,abs-2304-01373,abs-2304-03208} to design our proposed JIANG. The model is trained on a large quantity of textual data including both English and Chinese and use a standard optimizer.

\subsection{Pre-training Data}

Our training dataset is a mixture of several sources, covering Chinese materials, English materials, foundational knowledge, financial data, code, and conversations. For the most part, we reuse publicly available data sources. For some of the data, we built it from scratch.\\

\noindent \textbf{Chinese Internet [203.95B].} Using web crawling techniques, we extracted the titles and content information from certain web pages obtained from the Chinese Internet, by parsing the raw HTML of different websites. \\

\noindent \textbf{Wikipedia [24.03B].} We include Wikipedia dumps in both Chinese and English languages. 
We performed keyword filtering on the Chinese data, retaining the titles and content of knowledge-related entries. \\

\noindent \textbf{ThePile [82.77B].} Incorporating more English foundational knowledge into the model enables effective testing of bilingual capabilities. We thus include the publicly available ThePile dataset \cite{pile} in our data. \\

\noindent \textbf{GitHub [87.54B].} Some studies suggest that the use of code data during training process may be one of the sources for large language models to acquire logical reasoning abilities \cite{DBLP:conf/emnlp/MadaanZ0YN22}. Additionally, structured code logic may prove to be more helpful for future downstream tasks such as N2SQL, SQL2N, and others. We use the public GitHub dataset available on Hugging Face \footnote{\url{https://huggingface.co/datasets/codeparrot/github-code-clean}}. \\

\begin{table}[h]
\small
\centering
\caption{Pre-training data. Data mixtures used for pretraining, for each subset we list the sampling proportion and size.}
\label{tab:time}
\setlength\tabcolsep{1pt}
\begin{tabular}{ccc}
\toprule 
Dataset & Sampling prop. & Quantity(tokens) \\ 
\midrule
Chinese Internet    & 43.68\%   & 203.95 B  \\ 
Wikipedia       & 5.15\%  &  24.03 B \\ 
ThePile     & 17.73\%   &  82.77 B\\ 
GitHub    &  18.76\%  & 87.54 B \\ 
CLCF    &  9.63\%  & 44.94 B\\ 
Business Research Reports    & 3.54\%   & 16.55 B \\ 
ULCF   & 1.38\%   & 6.42 B \\ 
LCCC    & 0.11\%   & 0.52 B \\ 
\bottomrule 
\end{tabular}
\end{table}

\noindent \textbf{Chinese Listed Company Filling (CLCF) [44.94B].} A significant amount of open financial data serves as high-quality Chinese language corpora. We include a dump of Cninfo \footnote{\url{https://www.cninfo.com.cn}}, a large securities specialized website that provides public announcements and market data for listed companies. We parsed the original PDF documents of historical announcements from listed companies into plain text format, which served as the raw data.\\

\noindent \textbf{Business Research Reports [16.55B].} We collected a vast number of professional business research reports from the Internet. \\


\noindent \textbf{USA Listed Company Filling (ULCF) [6.42B].} We include a data dump of the SEC \footnote{\url{https://www.sec.gov}}, which contains quarterly and annual reports as well as other periodic filings submitted by publicly traded companies in the United States.\\

\noindent \textbf{LCCC [0.52B].} We include a publicly available large Chinese dialogue corpus originate from Chinese social medias on Hugging Face \footnote{\url{https://huggingface.co/datasets/lccc}}. \\

%
\noindent \textbf{Document Processing.} A rigorous data cleaning pipeline is designed to ensure the quality of the corpus. Low-quality documents such as too short, too long without punctuation, containing more than three NSFW (Not Safe for Work) terms, and fewer than 20 English words and Chinese characters are filtered. 

We utilize a multi-language Sentence-BERT model \cite{reimers-2019-sentence-bert} to vectorize the filtered documents and perform document extraction to enhance corpus diversity and achieve semantic deduplication. Specifically, we randomly select a document from a particular category and add it to the candidate pool. Then, we compute the semantic similarity of other documents with the ones in the candidate pool. From the 10\% of documents with the lowest similarity scores, we randomly select a new document and add it to the candidate pool. This process is repeated until a sufficient number of documents are chosen within the current category.\\

\noindent \textbf{BPE Tokenizer.}
We incorporated the basic tokenizer from GPT NeoX as a starting point. To better align with the code data, we enhanced the tokenizer of GPT NeoX by including corresponding embeddings. Additionally, JIANG introduced supplementary vocabulary in their tokenizer to accommodate Chinese words, encompassing a comprehensive vocabulary of commonly used Chinese characters. Based on our experimental findings, the training set exhibited a remarkable coverage rate of over 99.9\% for Chinese characters. JIANG's decision to utilize a tokenizer that covers individual characters rather than phrases like MOSS models primarily aims to avoid potential issues stemming from word segmentation. Furthermore, this approach facilitates future research possibilities, such as transitioning from the decoder to the encoder. Importantly, practical experience demonstrates that the choice of tokenizer does not significantly impact performance. For instance, GPT4 continues to employ a BPE tokenizer with a compact vocabulary, while achieving excellent results in Chinese text generation.

\subsection{Architecture}

\begin{table*}[t]
\centering
\caption{Performance Testing of JIANG Chinese Large-Scale Model and Other Existing Models.}
\label{tab:LLM-evaluation}
\aboverulesep=0ex
\belowrulesep=0ex
\begin{tabular}{ccccc}
\toprule 
task & aquila-7b-base & baichuan-7B-base & JIANG-base-65600 &moss-13B-base\\ 
\midrule
afqmc    & 0.3510 	&0.3123 	&0.4164 	&\textbf{0.5130} \\ 
boolq	&0.4612 	&\textbf{0.6835} 	&0.5905 	&0.6612  \\ 
examqa	&0.3517 	&\textbf{0.3707 }	&0.3410 	&0.3544 \\ 
iflytek	  &0.2097 	&0.2301 	&\textbf{0.2751} 	&0.2636 \\ 
climi	&0.3468 	&0.3125 	&0.3319 	&\textbf{0.3687} \\ 
cmnli	&0.7232 	&0.4945 	&0.7509 	&\textbf{0.9638}  \\ 
ocnli	&0.2644 	&0.2519 	&\textbf{0.2705} 	&0.2488 \\ 
pawsx\_zh	 &0.5265 	&0.5275 	&\textbf{0.5345} 	&0.4790 \\ 
\bottomrule 
\end{tabular}
\end{table*}

Incorporating insights from recent advancements in large language models, the design our network is based on the transformer architecture. Here, we highlight the key distinctions from the previous large language models and identify the sources of inspiration for these modifications:

\noindent  \textbf{Partially Cancellation of the Bias in Fully-connected Layer.} 
In the design of JIANG, only the bias value for calculating attention between query, key, and value is retained to improve the model's extrapolation capability. Additionally, the bias term in the remaining fully-connected layers has been removed.

\noindent \textbf{RMSNorm.} 
Drawing inspiration from prior research, RMSNorm layer is adopted in the model of JIANG. This choice is motivated by the fact that RMS Norm yields comparable performance to layer norm, yet significantly reduces computational overhead.

\noindent \textbf{Gated Mechanism.}
We have built upon the mechanism inherited from LLaMA, but with a distinct approach - we have implemented a more focused design for the model structure. Specifically, we have enhanced the layer count of the model, resulting in improved performance based on our experimental findings. These results will be detailed in the subsequent section.

\noindent \textbf{RoPE.}
We have chosen to implement RoPE \cite{/abs-2104-09864} as the position embedding approach in JIANG's model, building upon prior research. RoPE employs absolute position encoding to achieve relative position encoding, enhancing the model's extrapolation capability.

\noindent \textbf{FlashAttention.} Building on previous research, we have implemented FlashAttention \cite{DaoFERR22} to minimize the need for frequent reading and writing of the attention matrix in High Bandwidth Memory, thereby effectively reducing memory costs.

\subsection{Model Training}

During the training process, we employed a large batch size of 6 million tokens to enhance the model's stability. Initially, the sequence length was set to 2048, but it was increased to 4096 once the size of the trained data reached 100 billion. To optimize training efficiency, we leveraged various techniques such as deepspeed zero3, cpu offload, and data parallel processing. The training was conducted using 96 A100 80G GPUs, and the entire process took approximately 52 days.

\begin{figure}[t]
    \centering
    \includegraphics[width=0.5\textwidth]{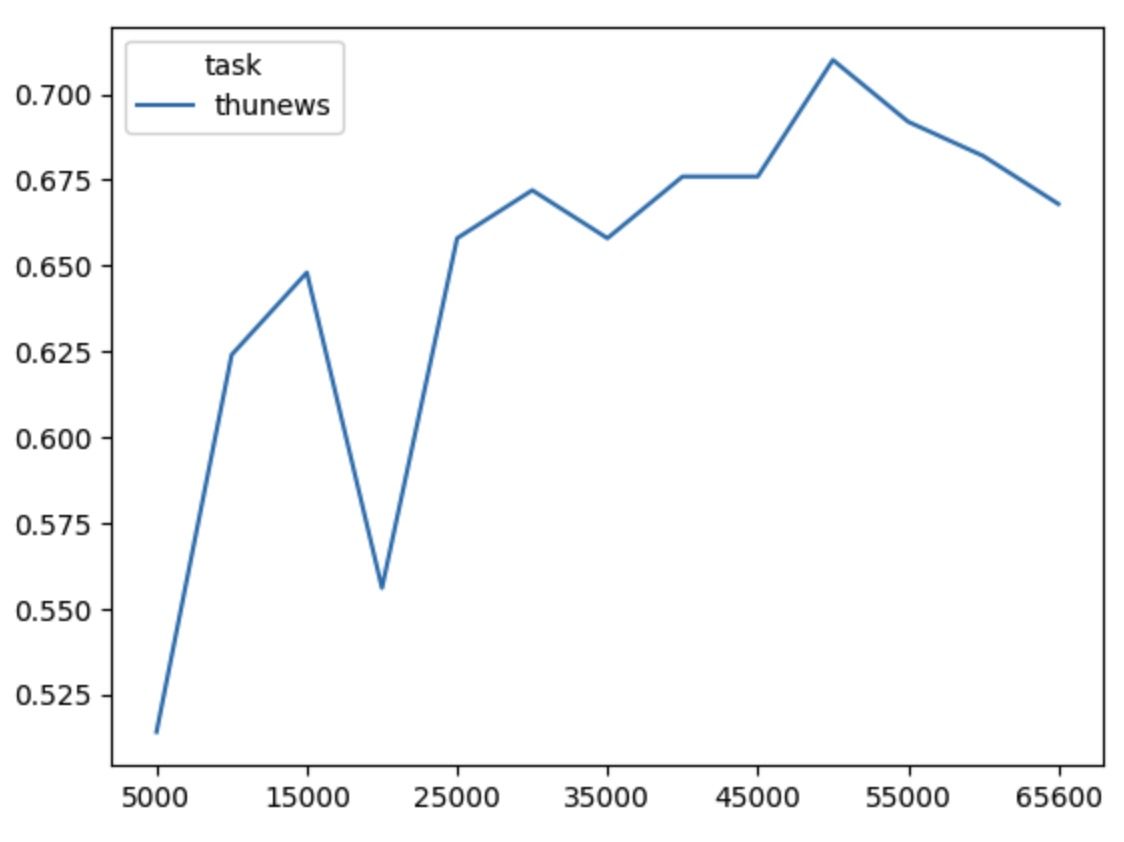}
    \caption{Model evaluation on thunews Dataset .}
    \vspace{-0.2cm}
	\label{fig:training}
\end{figure}
We also conducted model training testing by extracting 1000 questions from the thunews dataset. The model's performance was evaluated at every 5000 training steps. As illustrated in the graph, it is evident that the model's training exhibits a steady improvement over time.
\section{Experiments}

\begin{figure}[t]
    \centering
    \includegraphics[width=0.5\textwidth]{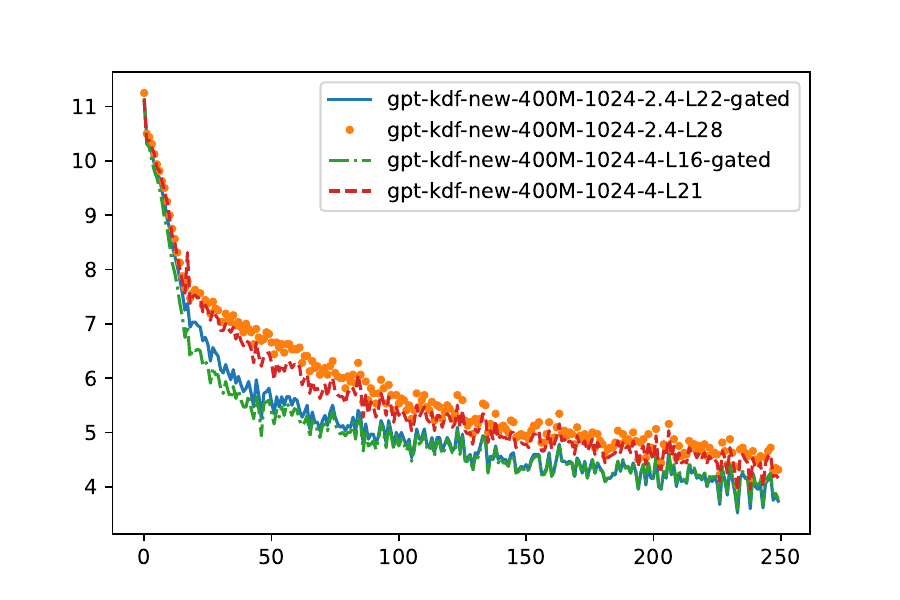}
    \caption{Model Framework Selection and Evaluation.}
    \vspace{-0.2cm}
	\label{fig:architecture}
\end{figure}
Extensive testing was conducted to refine the model framework design. Fig.~\ref{fig:architecture} presents the evaluation results of different models with a parameter size of 400 million, focusing on their convergence speed and effectiveness. The variations among the four models depicted in the graph primarily stem from the integration of the Gated Mechanism and the parameter configuration of the fully connected layer. Notably, the model represented by the blue line, labeled gpt-kdf-new-400M-1024-2.4-L22-gated, demonstrated superior performance, achieving the fastest convergence speed.

In our study, we examined the performance of JIANG Chinese large-scale model and other prominent models across eight tasks, adhering to LM-Eval standards\footnote{\url{https://github.com/EleutherAI/lm-evaluation-harness}}.table.~\ref{tab:LLM-evaluation} illustrates the results of our testing. We observed that JIANG Chinese model performed slightly lower than other models in English tasks like boolq and examqa. We attribute this difference to the relatively smaller English language corpus used during training, which limited the model's English reasoning capabilities. However, in Chinese tasks such as iflytek, ocnili, and pawsx\_zh , JIANG Chinese model achieved the highest scores among the four models. This indicates that our JIANG model, with its specific architecture and training on Chinese corpora, demonstrates commendable performance in Chinese reasoning. In addition, in the cmnli task, MOSS achieved an impressive score of 0.9638, whereas our JIANG model attained a score of 0.7509. Indeed, our JIANG model's performance is second only to MOSS, but the difference of 0.2 points indicates that there is room for improvement in certain areas. This highlights the potential for further enhancements in our model.Overall, when compared to similar Chinese models like MOSS, our JIANG Chinese model demonstrates comparable capabilities and ranks among the top-tier of Chinese large-scale models. We firmly believe that the emergence and open-sourcing of the JIANG Chinese model will further accelerate the advancement of Chinese large-scale models.
\section{Conclusion}

In this paper, we present JIANG, our proposed Large Language Model specifically tailored for the Chinese language. We introduce the data collection process and the model's design, and showcase the exceptional performance of JIANG in various natural language processing tasks in Chinese.

\bibliographystyle{ACM-Reference-Format}
\bibliography{custom}

\end{document}